\documentclass[runningheads]{llncs}
\usepackage{orcidlink}
\renewcommand{\orcidID}[1]{\,\orcidlink{#1}}
\usepackage[T1]{fontenc}
\usepackage{graphicx}
\usepackage{booktabs}
\usepackage{geometry}
\begin{document}
%
\title{Benchmarking Off-the-Shelf Multimodal AI Models Against Dermatologists on Patient-Captured Skin Images}
\titlerunning{Benchmarking AI Models on Dermatological Images}
%
\author{Rian Dolphin\inst{1}\orcidID{0000-0002-5607-9948} \and
Laura Knowles\inst{2}\orcidID{0009-0007-1143-2107}}

\authorrunning{R. Dolphin \& L. Knowles}

\institute{
Independent Research, Dublin, Ireland \\
\and
School of Medicine, University of Limerick, Ireland
}

\maketitle              
\begin{abstract}
Artificial intelligence (AI) has advanced at a rapid pace in recent years. Initially, breakthroughs in large language models caught widespread attention. However, recent generations of frontier AI models have adopted multimodal capabilities as a first class citizen, with vision capabilities being central to that. In this paper, we evaluate three recently released models on the task of diagnosing dermatological conditions from patient-submitted images. The models chosen are at the low to mid tier in terms of pricing and thus represent a floor on current AI capabilities, not a ceiling. We evaluate AI performance relative to a panel of three certified dermatologists, who grade each image, and we present four interesting findings. Firstly, depending on the metric, the tested AI models are either on par or slightly trail humans in terms of inter-clinician agreement. Secondly, we find that asking AI models for a confidence rating produces poorly calibrated answers, meaning use of confidence thresholds should not be relied upon in a clinical setting. Thirdly, the effect of providing additional patient metadata is strongly model-specific, with one of the three models degrading on every metric considered. Finally, model cost is not predictive of performance. The best-performing model we tested costs on average \$0.0045 per case.

\end{abstract}

\section{Introduction}

Deep learning has been applied to medical imaging for over a decade, typically in the form of task-specific models trained on curated datasets for a single condition or modality. More recently, a new class of generalist vision language models has emerged. Unlike their task-specific predecessors, these models are not trained for any particular clinical task; they can instead be given an image together with a natural-language prompt and return a structured response through a single API request. A natural question is how well such generalist models perform when applied to real-world clinical tasks and datasets.

Dermatology is a natural domain in which to pose this question. A common first step in dermatological care involves a patient capturing a photograph of a skin concern before deciding whether to seek an in-person clinical visit. A system that can accurately produce a differential diagnosis from such an image therefore has obvious applications in clinical triage. Patient-submitted images are, however, a challenging input distribution for automated systems, as they are typically captured under variable lighting, framing, and image quality conditions \cite{rikhye2024differences}. Whether a generalist vision language model can produce differential diagnoses from such images that are consistent with those of a certified dermatologist is therefore an open empirical question, and is the focus of this paper.

\begin{figure}[tb]
\centering
\includegraphics[width=\linewidth]{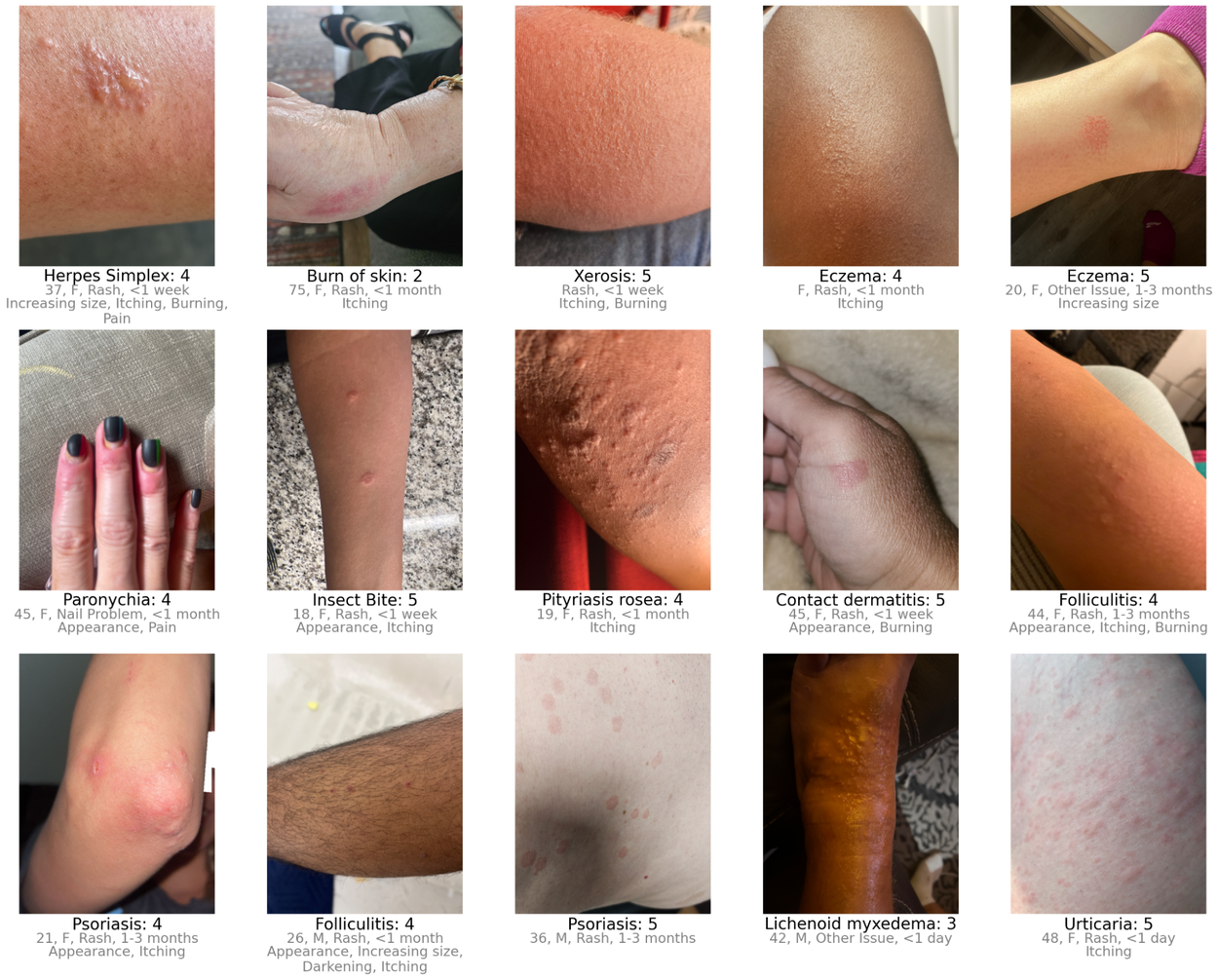}
\caption{A sample of sixteen labelled cases from the SCIN release, illustrating the breadth of conditions and Fitzpatrick skin types in the dataset. The images reflect real phone-camera conditions rather than clinical dermatoscopy: mixed household lighting, variable framing, and typical hand-held distances. Reproduced from Ward et al.~\cite{ward2024creating}.}
\label{fig:label_examples}
\end{figure}

In this paper we evaluate three recently released vision language models on the task of diagnosing dermatological conditions from patient-submitted images. The images are drawn from the SCIN dataset~\cite{ward2024creating}, a crowdsourced collection of dermatology cases in which each case is reviewed by up to three certified dermatologists. Figure~\ref{fig:label_examples} shows a sample of sixteen such cases, illustrating the breadth of conditions and skin types present in the dataset. The three models we evaluate sit at the low to mid tier in terms of pricing. The top-tier flagship models were not included due to the cost of running the full experiment against them, meaning our results should be read as a floor on current AI capabilities rather than a ceiling. To ensure a fair comparison between AI and clinicians, we adopt a leave-one-out protocol. For each case, one of the three dermatologists is held out and the remaining two are used to form a peer reference. Both the held-out dermatologist and each AI model are then graded against the same peer reference, on the same cases, and with the same number of diagnoses permitted.

From this comparison we present four findings. Firstly, depending on the metric considered, the tested AI models are either on par with or slightly trail an individual US board certified dermatologist in terms of inter-clinician agreement. Secondly, we find that asking the models to report a confidence rating produces poorly calibrated answers, and that the reported confidence at most weakly tracks case difficulty, meaning confidence scores should not be relied upon in a clinical setting. Thirdly, model cost is not predictive of performance, with the cheapest model in our set producing the best results on every metric considered. Finally, the effect of providing additional patient metadata such as age, body site, symptoms, and duration is strongly model-specific: one model improves on every metric when the metadata is supplied, while another degrades on every metric. The latter is a somewhat counterintuitive finding, as one might expect additional patient context to improve a model's diagnosis, yet for one of the three models the opposite is observed throughout.

\section{Related Work}

Automated dermatological diagnosis has historically been approached through task-specific deep learning. Esteva et al.~\cite{esteva2017dermatologist} trained a convolutional neural network on roughly 130{,}000 clinical images and reported dermatologist-level performance on the binary classification of keratinocyte carcinomas and melanomas, and Liu et al.~\cite{liu2020deep} extended this line of work to a deep learning system for differential diagnosis across a much broader space of skin conditions. Subsequent audits of this generation of models identified systematic performance gaps across Fitzpatrick skin types and contributor populations~\cite{daneshjou2022disparities,groh2021evaluating}, motivating the release of dermatology datasets that are explicitly unenriched for condition category and skin type. The SCIN dataset~\cite{ward2024creating} used in this paper is one such release, and its labelling protocol, in which each case is reviewed by up to three dermatologists who each return up to three ranked diagnoses with a five-point confidence rating, is inherited from the labelling protocol used by Liu et al.~\cite{liu2020deep}. Our study reuses SCIN and its labelling conventions, but replaces the specialist classifier at the centre of that pipeline with a commercially available generalist vision language model, and evaluates the result under a matched leave-one-out protocol against an individual dermatologist.

A second line of work reframes medical image understanding as a capability of general-purpose multimodal foundation models rather than of task-specific classifiers. Moor et al.~\cite{moor2023foundation} set out the generalist medical AI paradigm, in which a single multimodal model is prompted at inference time for many clinical tasks rather than trained from scratch for each one. Several purpose-built medical vision language models follow this paradigm, including Med-PaLM-M~\cite{tu2024towards} and Med-Gemini~\cite{yang2024advancing}, both of which are evaluated on dermatology among other modalities. These models are trained or fine-tuned on curated biomedical corpora, and are not generally available through the same commercial APIs as their base models. We take a different position. We evaluate three off-the-shelf commercial generalist vision language models, none of which have received any disclosed medical fine-tuning, under a single API call per case on the same dermatology task. The question we ask is therefore not what ceiling a dedicated medical foundation model can reach, but how close a generic, commercially priced model already sits to an individual dermatologist when used as-is.

Two further threads of prior work are directly relevant to our non-accuracy findings. Kadavath et al.~\cite{kadavath2022language} show that large language models can produce reasonably calibrated token-level probabilities on multiple-choice benchmarks, but that self-reported probability of correctness judgements generalise only partially and degrade out of distribution. Our calibration result is consistent with the cautionary side of that picture. We elicit confidence through a prompted one-to-five rating rather than a token probability, and we find that the resulting ratings are both systematically overconfident relative to empirical dermatologist agreement and essentially uncorrelated with case difficulty, which makes them unsafe as a clinical gating signal. Separately, Rikhye et al.~\cite{rikhye2024differences} characterise the distribution shift between clinician-captured and patient-captured dermatology images and identify condition distribution, rather than image quality, as the primary source of generalisation error for dermatology AI. The SCIN cases we evaluate are on the patient-captured side of that shift, and the counterintuitive direction of our patient-metadata finding, in which two of three models regress when structured patient context is added, has, to our knowledge, not been reported in the prior task-specific literature.

\section{Data \& Experimental Setup}\label{sec:data}

Our experiment evaluates three recently released vision language models against a held-out dermatologist on a subset of the SCIN dataset. In this section we describe the dataset itself, the filtering applied to obtain our evaluation subset, and the prompting configuration under which each model is called. The leave-one-out comparison protocol and the agreement metrics used to grade both the models and the dermatologists are described separately in Section~\ref{sec:eval}.

\subsection{Dataset}

SCIN~\cite{ward2024creating} is a crowdsourced dermatology dataset containing 5{,}033 cases collected between March and November 2023 via targeted web-search advertisements. Each case consists of one to three photographs captured by the contributor, together with self-reported demographic information and symptom metadata. Each case is reviewed by between one and three US board-certified dermatologists drawn from a pool of ten, with a mean clinical experience of 14.8 years. Dermatologists provide up to three ranked differential diagnoses per case, each with an associated confidence rating on a five-point scale. The ranked labels from each dermatologist are aggregated into a weighted consensus using a rank-inverse weighting scheme, in which the top-ranked diagnosis receives weight one, the second weight one half, and the third weight one third; these weights are then summed across dermatologists and normalised to sum to one.

\begin{figure}[tb]
\centering
\includegraphics[width=0.9\linewidth]{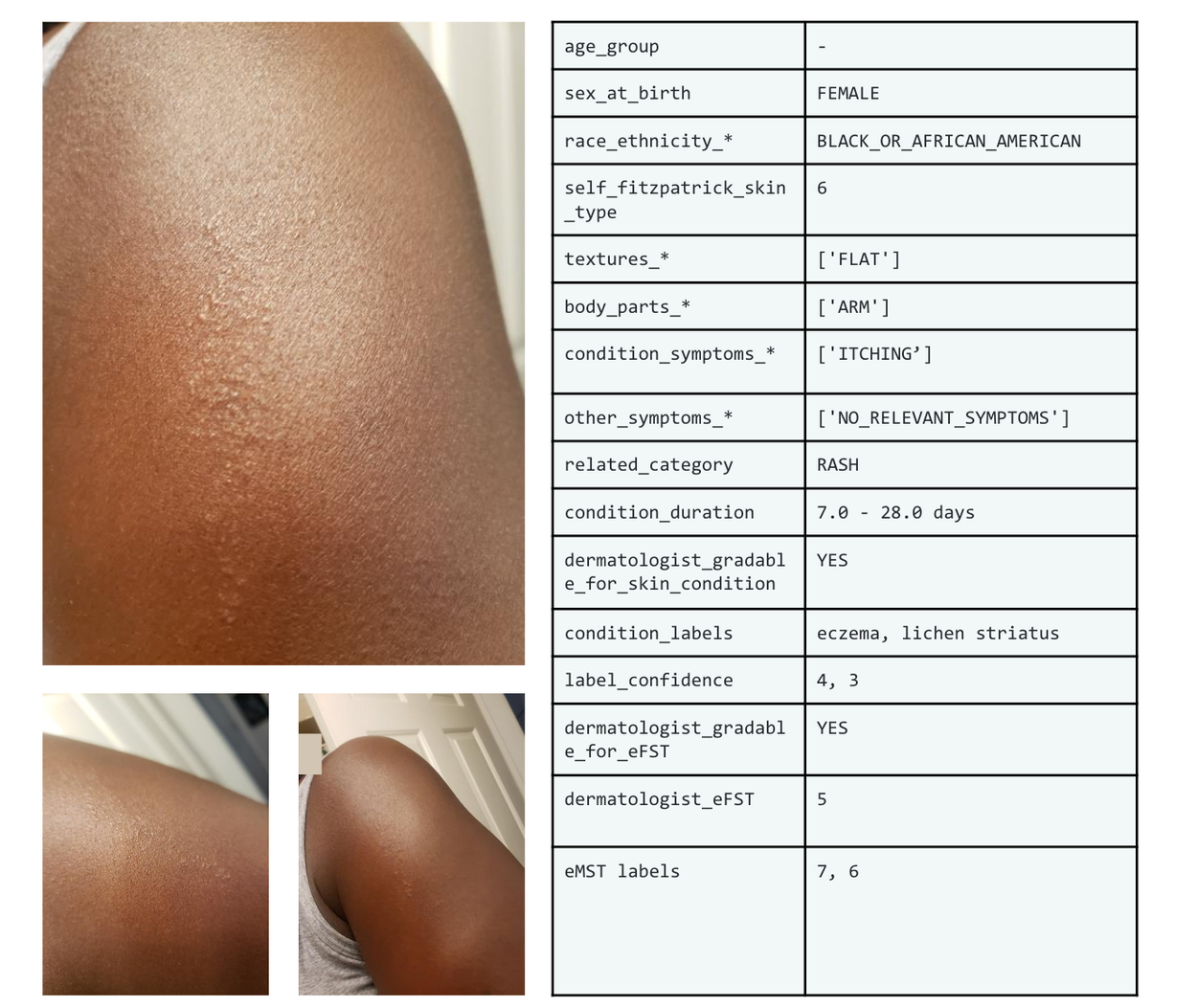}
\caption{An example SCIN case. Left: the patient-captured photographs submitted for the case. Right: the structured metadata stored for the same contributor, including age group, sex at birth, self-reported race/ethnicity and Fitzpatrick skin type, body part, symptoms, condition duration, and the per-dermatologist diagnoses and confidences. Reproduced from Ward et al.~\cite{ward2024creating}.}
\label{fig:example_case}
\end{figure}

Unlike earlier dermatology datasets that focus predominantly on skin cancer and are captured under controlled clinical conditions, SCIN is explicitly designed to be unenriched for condition category, skin type, or contributor population. It therefore contains a broad distribution of inflammatory and infectious conditions in addition to neoplastic ones, and covers a wider range of Fitzpatrick skin types than most publicly available dermatology datasets. Figure~\ref{fig:example_case} shows one such case, including the patient-captured photographs and the structured metadata that SCIN records for the same contributor.

\subsection{Evaluation Subset}

\begin{figure}[tb]
\centering
\includegraphics[width=\linewidth]{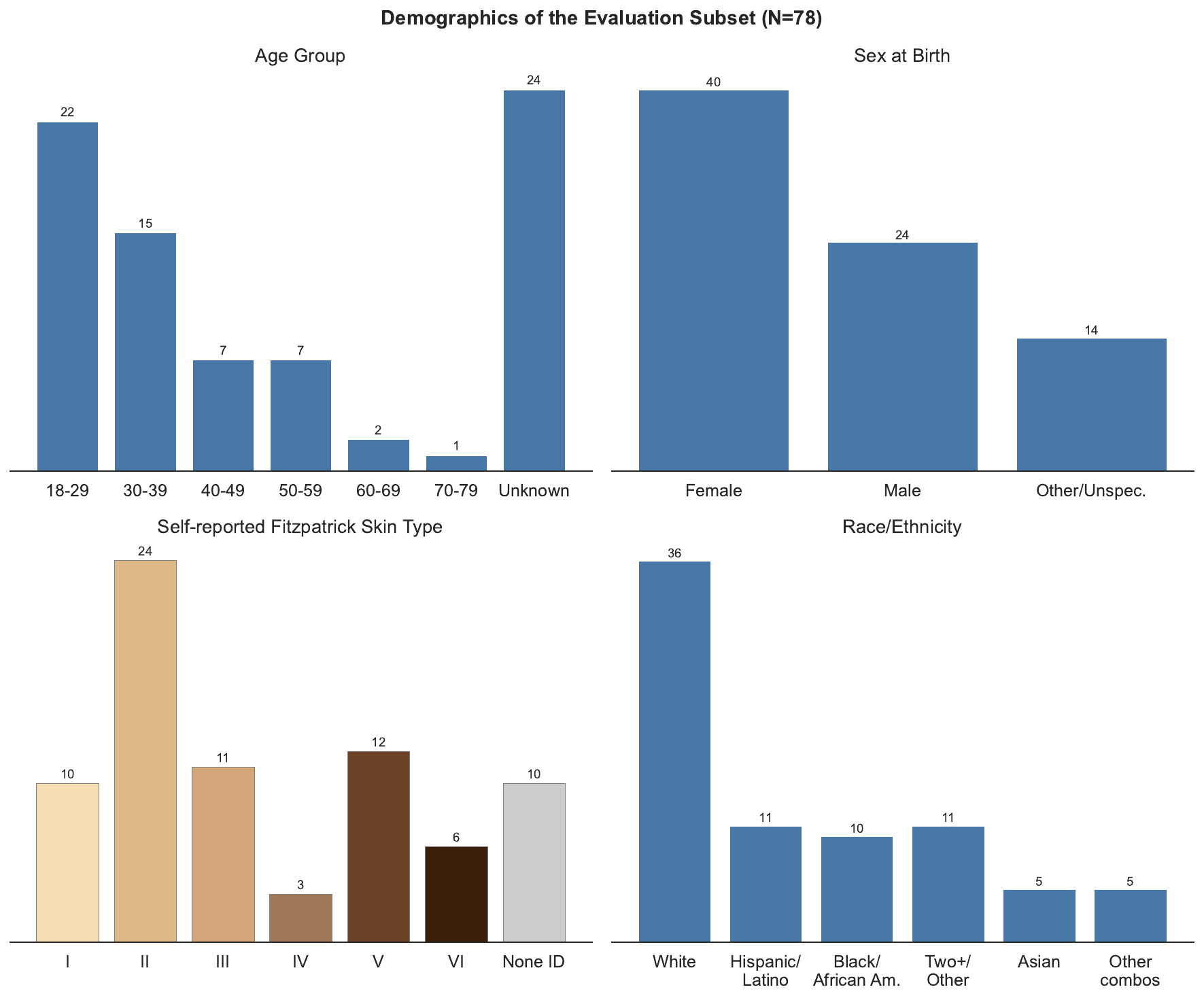}
\caption{Demographics of the 78 cases in our evaluation subset. Women contribute 51\% of cases, 41\% of known-age contributors are aged 18--29, and 23\% self-report Fitzpatrick skin type V or VI. Around 18\% of cases have unspecified sex and 31\% have unknown age, as SCIN collects self-reported demographics for only about half of contributors.}
\label{fig:demographics}
\end{figure}

Our evaluation requires each case to have been reviewed by all three dermatologists, so that a leave-one-out comparison between the models and an individual dermatologist is possible. The SCIN paper describes this triple-labelled cohort as the first 750 submissions with self-reported health and demographic information, and the public release contains 716 of these cases. Of these 716 cases, we begin with the 318 on which all three dermatologists independently marked the images as gradable.

A further complication arises from the structure of the labels in the public release. Although each of the 318 cases is known to have been reviewed by three dermatologists, the release concatenates all of their diagnoses into a single flat list with no explicit rater index, so the specific diagnoses attributed to each individual dermatologist cannot be recovered directly. To enable the leave-one-out comparison, we reconstruct the per-dermatologist groupings using the observation that each dermatologist lists their diagnoses in descending confidence order, meaning that a confidence increase within the flat list marks a new rater boundary. A reconstruction is considered valid when it produces exactly three non-empty groups of one to three diagnoses each, with no duplicate condition within any group. This procedure yields a valid reconstruction on 78 of the 318 gradable cases, and this set of 78 cases forms the evaluation subset used throughout the remainder of the paper.

This filtering is not uniform across the gradable set, and tends to retain cases on which the dermatologist panel was less decisive. The 78 cases in our evaluation subset contain more diagnoses per case on average, a larger number of unique conditions, lower mean dermatologist confidence, and a less peaked weighted consensus than the 240 cases that are excluded. The absolute agreement numbers we report should therefore be interpreted as measured on comparatively harder cases, and applied to an easier sample the absolute agreement numbers for both the models and the dermatologists would likely be higher. Figure~\ref{fig:demographics} shows the demographic composition of the 78-case evaluation subset.

\subsection{Models and Prompting}

We evaluate three vision language models, one from each of OpenAI, Anthropic, and Google: GPT-4o-mini, Claude Sonnet 4.6, and Gemini 3 Flash Preview. All three models are called through a common wrapper built on the Pydantic-AI library, with an identical system prompt and an identical structured output schema across models. Each model is instructed to act as a board-certified dermatologist, to first assess whether the image is of sufficient quality to grade, and to then return up to three ranked differential diagnoses. Each diagnosis consists of a condition name, a confidence rating on the same one-to-five scale used by the SCIN dermatologists, and a short clinical rationale.

Each model is evaluated under two prompt conditions. In condition A, the model is supplied only with the patient-captured photographs for the case. In condition B, the model is additionally supplied with the patient-reported metadata available in the SCIN release, including age group, sex at birth, self-reported Fitzpatrick skin type, body location, symptoms, condition duration, and the patient-reported condition category. The comparison between conditions A and B allows us to isolate the effect of supplying structured patient context alongside the images on model performance.

Because the models return free-text condition names that do not always correspond directly to the 370 condition categories used by SCIN, we map each predicted condition to a SCIN category using a deterministic pipeline consisting of manual synonyms, exact matching, fuzzy matching, and substring containment. Any predicted condition that cannot be mapped to a SCIN category is treated as incorrect on every agreement metric used in our evaluation. Mapping coverage under condition A is 100\% for GPT-4o-mini and Gemini 3 Flash Preview, and 89.4\% for Claude Sonnet 4.6, which tends to produce more specific diagnostic phrasings that do not always route to one of the SCIN categories. The implications of this coverage asymmetry for the absolute agreement numbers are discussed in Section~\ref{sec:limitations}.

\section{Evaluation}\label{sec:eval}

In this section we describe how the model and dermatologist predictions are compared against one another. Our guiding principle is that every number we report should be directly comparable between the two: the same indicator function is applied to both a model and a dermatologist, both are evaluated against the same peer reference, and both are allowed the same number of diagnoses at each position. Any difference between a model's agreement rate and the dermatologist baseline can therefore be attributed to the predictions themselves, rather than to asymmetries in how each side is graded.

\subsection{Leave-One-Out Protocol}

For each of the 78 cases in our evaluation subset we cycle through the three dermatologists in turn. At each position, one dermatologist is held out and the remaining two are used to form a peer reference. A peer consensus ranking is then computed from the two remaining dermatologists using the same rank-inverse weighting scheme that SCIN applies across three dermatologists, in which the top-ranked diagnosis of each dermatologist contributes weight one, the second contributes weight one half, and the third contributes weight one third. These weights are summed across the two peer dermatologists and normalised to produce the two-dermatologist peer reference for that held-out position.

Both predictors are then evaluated against this same peer reference: the held-out dermatologist's top-$k$ differential, and the model's top-$k$ predictions. Here $k$ denotes the number of diagnoses listed by the held-out dermatologist at that position, which may be one, two, or three depending on the dermatologist and the case. The model's predictions are capped to $k$ so that both the held-out dermatologist and the model contribute the same number of diagnoses at each position. Across the evaluation subset, 31\% of held-out positions have $k = 1$, 42\% have $k = 2$, and 26\% have $k = 3$. Every metric we report is averaged over the $3 \times 78 = 234$ held-out positions obtained in this way.

\subsection{Agreement Metrics}

We report four agreement metrics, each measuring a different level of strictness at which the predictor can be said to have agreed with the two-dermatologist peer reference. The metrics vary along two dimensions: whether the predictor is restricted to its top-1 prediction or allowed up to $k$ attempts, and whether a match requires agreement from only one of the two peer dermatologists or from both.

\begin{itemize}

    \item \textit{Q1: Any overlap.} Did the predictor's top-$k$ differential contain any condition listed by at least one of the two peer dermatologists? This is the most relaxed of the four metrics and asks simply whether the predictor reached for something that a peer also reached for.

    \item \textit{Q2: Top-1 in peer bag.} Is the predictor's top-1 prediction a condition listed by at least one of the two peer dermatologists? This is stricter than Q1 because the predictor is given only a single attempt rather than $k$.

    \item \textit{Q3: Top-$k$ with peer agreement.} Did the predictor's top-$k$ differential contain any condition listed by both peer dermatologists? This is stricter than Q1 because the matched condition must have been listed by both peers, rather than by only one.

    \item \textit{Q4: Consensus match.} Is the predictor's top-1 prediction the same as the top-ranked diagnosis of the two-dermatologist peer consensus, computed using the rank-inverse weighting scheme described above? This is the strictest of the four metrics, combining the top-1 restriction of Q2 with the both-peers-agree requirement of Q3.

\end{itemize}

Q1 is therefore the most relaxed of the four metrics and Q4 is the strictest. Q2 and Q3 each make Q1 stricter along one of the two dimensions, but neither is directly stricter than the other.

\section{Results}

\begin{table}[tb]
\centering
\caption{Agreement metrics Q1--Q4 on our 78-case evaluation subset. Each number is the mean over the $3 \times 78 = 234$ held-out positions. The column labelled ``LOO baseline'' (\textbf{L}eave \textbf{O}ne \textbf{O}ut) gives the agreement rate obtained when the predictor is the held-out human dermatologist, and is directly comparable with the model columns in each row. Under each model, column A corresponds to the images-only prompt condition and column B corresponds to the condition in which patient metadata is additionally supplied. The best model cell in each row is shown in bold.}
\label{tab:primary}
\begin{tabular}{l c c c c c c c}
\toprule
Metric & LOO & \multicolumn{2}{c}{Gemini 3 Flash} & \multicolumn{2}{c}{Claude Sonnet 4.6} & \multicolumn{2}{c}{GPT-4o-mini} \\
\cmidrule(lr){3-4}\cmidrule(lr){5-6}\cmidrule(lr){7-8}
 & baseline & A & B & A & B & A & B \\
\midrule
Q1: any overlap              & 0.744 & \textbf{0.684} & 0.645 & 0.568 & 0.641 & 0.551 & 0.530 \\
Q2: top-1 in peer bag        & 0.581 & \textbf{0.551} & 0.506 & 0.432 & 0.500 & 0.342 & 0.393 \\
Q3: top-$k$ with both peers  & 0.308 & \textbf{0.308} & 0.273 & 0.248 & 0.261 & 0.192 & 0.205 \\
Q4: top-1 $=$ peer consensus & 0.286 & \textbf{0.291} & 0.251 & 0.218 & 0.226 & 0.171 & 0.222 \\
\bottomrule
\end{tabular}
\end{table}

We now present the results of the leave-one-out comparison described in the previous section. We first report the four agreement metrics on our evaluation subset, comparing each model to the individual-dermatologist baseline. We then examine how each model's agreement changes when patient metadata is added to the prompt, and finally we assess how well each model's self-reported confidence ratings track its actual agreement with the dermatologist panel.

\subsection{Agreement with Dermatologists}
\label{sec:primary}

\begin{figure}[tb]
\centering
\includegraphics[width=\linewidth]{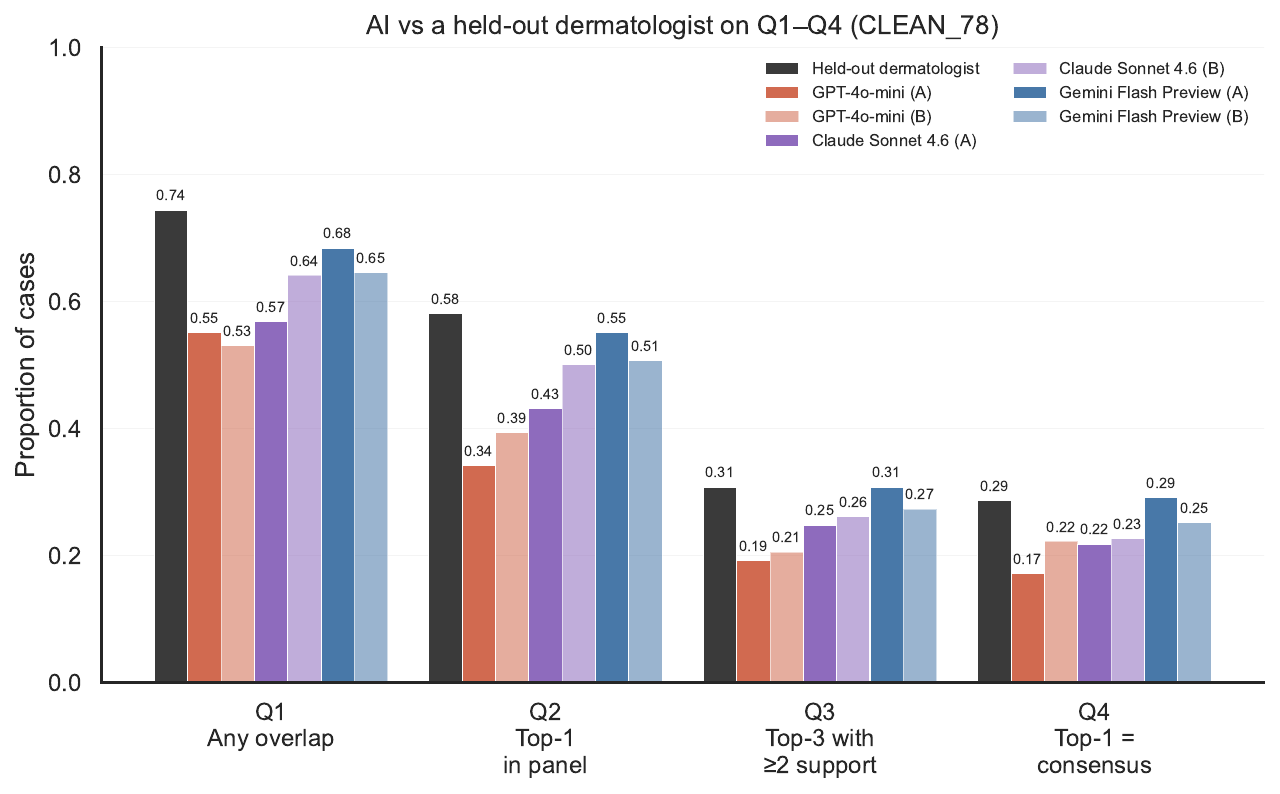}
\caption{The four agreement metrics computed on our evaluation subset. The grey bar in each group corresponds to the individual-dermatologist baseline. The coloured bars correspond to the three models, shown separately for each of the two prompt conditions (A: images only; B: images plus patient metadata).}
\label{fig:primary}
\end{figure}

Table~\ref{tab:primary} and Figure~\ref{fig:primary} present the four agreement metrics under both prompt conditions, together with the individual-dermatologist baseline obtained by applying the same leave-one-out protocol to a held-out dermatologist. Because the peer reference and the number of allowed diagnoses are identical across the two predictors at each held-out position, each row of the table is directly comparable.

Several observations follow from the table. First, the rank order among the three models is consistent across all four metrics and both prompt conditions, with Gemini 3 Flash Preview producing the highest agreement on every cell, followed by Claude Sonnet 4.6, and then GPT-4o-mini. This ordering is notable in that Claude Sonnet 4.6 is approximately 5$\times$ the price of Gemini 3 Flash Preview, which indicates that within the mid-tier price range considered in our study, model cost is not predictive of performance on this task.

Second, the gap between the best-performing model and the individual-dermatologist baseline depends on which metric is considered. On the two strictest metrics, Q3 and Q4, Gemini Flash Preview without any patient metadata (condition A) reaches agreement rates that essentially match the individual-dermatologist baseline. On Q4, the baseline is 28.6\% and Gemini 3 Flash reaches 29.1\%. On Q3, the baseline is 30.8\% and Gemini 3 Flash reaches exactly 30.8\%. On the two more relaxed metrics, however, all three models trail the baseline. On Q1, which asks whether any prediction appears in either peer's label bag, the baseline is 74.4\% and the best model reaches 68.4\%, a gap of approximately six percentage points. On Q2, which asks whether the top-1 prediction appears in either peer's label bag, the baseline is 58.1\% and the best model reaches 55.1\%. The overall pattern is that the tested models are essentially at parity with an individual dermatologist on the strictest agreement questions, but several percentage points behind on the more relaxed ones.

We note one caveat relating to the absolute numbers reported for Claude Sonnet 4.6. As described in Section~\ref{sec:data}, the mapping coverage for Claude Sonnet 4.6 under condition A is 89.4\%, compared to 100\% for the other two models, because Claude tends to return more specific diagnostic phrasings that do not always route to a SCIN category. Any unmapped prediction is treated as incorrect on every metric. We estimate that this depresses Claude Sonnet 4.6's absolute Q4 number by approximately five percentage points. The rank order among the three models is robust to this asymmetry, but the absolute gap between Claude Sonnet 4.6 and Gemini 3 Flash Preview may be overstated. The implications of the mapping asymmetry are discussed further in Section~\ref{sec:limitations}.

\subsection{Effect of Patient Metadata}
\label{sec:metadata}

\begin{figure}[tb]
\centering
\includegraphics[width=1\linewidth]{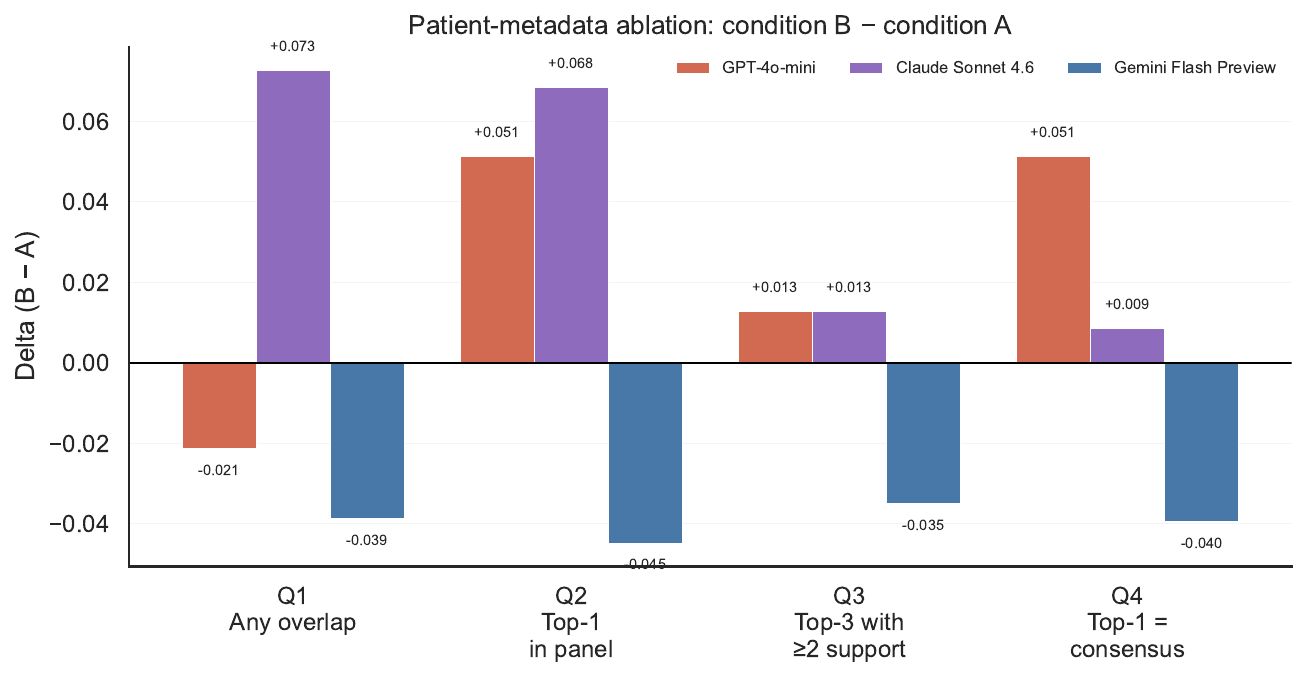}
\caption{Change in each agreement metric when patient metadata (age, sex at birth, self-reported Fitzpatrick skin type, body site, symptoms, condition duration, and patient-reported condition category) is added to the prompt. Positive bars indicate that the model's agreement improved when the metadata was supplied; negative bars indicate that it regressed.}
\label{fig:metadata}
\end{figure}

Figure~\ref{fig:metadata} shows the change in each agreement metric when the prompt is augmented with patient-reported metadata in condition B, relative to the images-only condition A.

Of the three models evaluated, only Claude Sonnet 4.6 shows consistent improvement when patient metadata is added to the prompt. It gains 7.3 percentage points on Q1, 6.8 percentage points on Q2, and smaller positive changes on Q3 and Q4. Gemini 3 Flash Preview regresses on all four metrics when metadata is added to the prompt, with losses of 3.9 percentage points on Q1, 4.5 on Q2, 3.5 on Q3, and 4.0 on Q4. GPT-4o-mini improves on Q2, Q3, and Q4 and regresses on Q1.

The overall pattern is that the effect of supplying structured patient context is strongly model-specific. We do not speculate on the underlying mechanism, but we note that the three models are called through an identical wrapper with an identical prompt and an identical structured output schema, so the differing responses are driven by the models themselves rather than by differences in how the metadata is presented to each.

\subsection{Confidence Calibration}
\label{sec:calibration}

\begin{figure}[tb]
\centering
\includegraphics[width=1\linewidth]{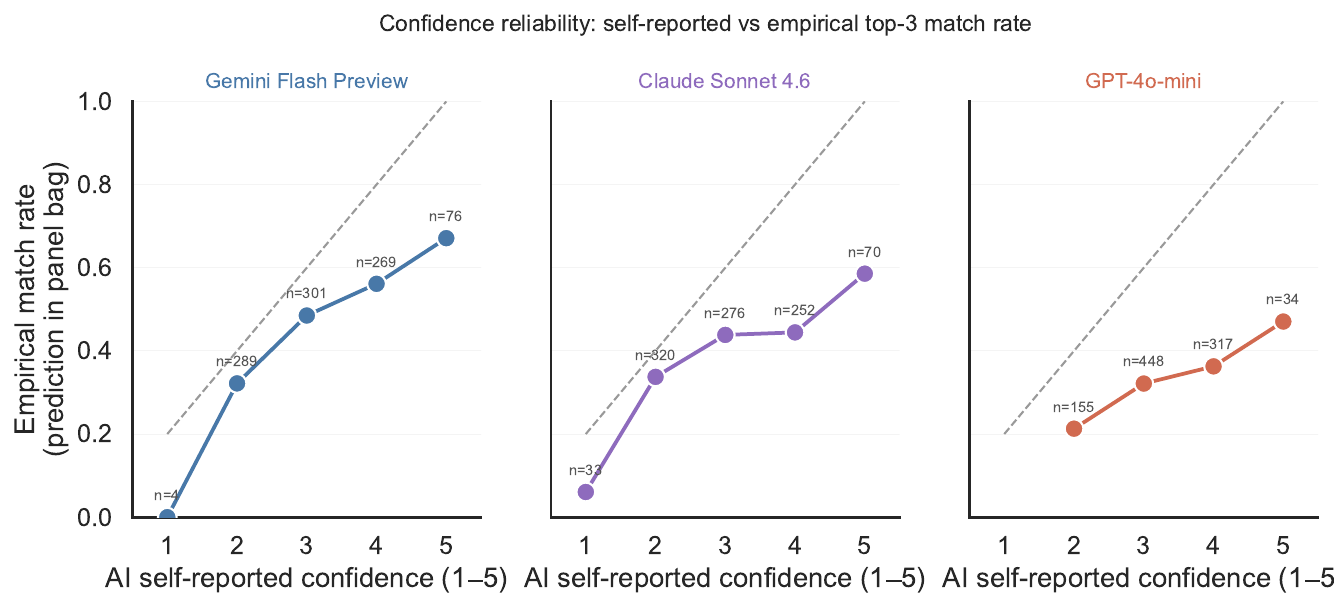}
\caption{Reliability diagrams for the three models under condition A (images only), computed over all ranked diagnoses on the 318-case gradable subset. The horizontal axis is the model's self-reported confidence on a one-to-five scale; the vertical axis is the fraction of predictions at each confidence level whose mapped condition is listed by at least one of the three dermatologists who reviewed the case. The dashed line shows the diagonal that a perfectly calibrated confidence rating would follow.}
\label{fig:calibration}
\end{figure}

Figure~\ref{fig:calibration} plots the self-reported confidence of each model against the empirical rate at which its predictions are listed by at least one of the three dermatologists who reviewed the same case. Because this analysis does not require the per-dermatologist reconstruction, it is computed over all ranked diagnoses on the full 318-case gradable subset described in Section~\ref{sec:data}, which provides a larger sample at each confidence level than the 78-case evaluation subset. A perfectly calibrated confidence rating would lie on the diagonal dashed line, indicating that a self-reported confidence of $c$ out of five corresponds to an empirical agreement rate of $c/5$.

Two observations can be drawn from the reliability diagrams. First, all three models' self-reported confidence ratings are miscalibrated relative to the empirical dermatologist-agreement rates. The reliability curves are monotonic, meaning that a higher self-reported confidence does correspond to a higher empirical agreement rate, but all three curves sit below the diagonal, meaning that the models are systematically overconfident relative to how often they actually agree with the dermatologist panel. At the top confidence rating of five out of five, Gemini 3 Flash Preview achieves 67\% empirical agreement, Claude Sonnet 4.6 sits in the middle, and GPT-4o-mini achieves only 47\%. Pooled across all ranked diagnoses, the mean confidence reported by the models is 3.0 to 3.2 on the five-point scale, compared to 2.65 for the SCIN dermatologists on the same subset.

Second, and more consequentially for any clinical use of these models, the top-1 confidence rating of each model is nearly constant across cases. On the same 318-case subset, the mean top-1 confidence across models ranges from 4.1 to 4.2, with a standard deviation of only 0.3 to 0.5. More tellingly, the top-1 confidence at most weakly tracks case difficulty. Measuring difficulty as the mean top-1 confidence of the three dermatologists on the same case, the Pearson correlation is 0.05 for GPT-4o-mini, 0.08 for Gemini 3 Flash Preview, and 0.18 for Claude Sonnet 4.6. \textit{The models therefore report essentially the same confidence on a case regardless of whether the dermatologist panel found the case easy or difficult to grade.}

This has a direct clinical implication. A triage workflow that uses the model's top-1 confidence as a threshold for referral, for example referring a case to a human clinician whenever the model reports low confidence on its top-1 diagnosis, will not catch the cases on which the dermatologists themselves were uncertain, because the model reports essentially the same top-1 confidence on those cases as on any other. Confidence gating should therefore not be relied upon in a clinical setting when using these models.

\section{Limitations}\label{sec:limitations}

Our 78-case evaluation subset is small and is not representative of the full SCIN release. It is heuristically selected from the 318 cases in which all three dermatologists marked the images as gradable, and as discussed in Section~\ref{sec:data} it retains cases on which the panel was less decisive than average. The absolute agreement numbers we report should therefore not be extrapolated to the full SCIN dataset, and the sample size of 78 bounds the statistical precision of the reported rates. The per-dermatologist reconstruction used to enable the leave-one-out comparison is also heuristic, and can mis-attribute a diagnosis to the wrong dermatologist even in cases where the resulting split is protocol-valid.

Our two-dermatologist peer reference does not literally reproduce the three-dermatologist weighted consensus stored in the SCIN release, and differs from the stored consensus by up to 0.17 on sample cases. This most likely reflects a tie-breaking or deduplication rule that we were unable to recover from the SCIN documentation. The individual-dermatologist baseline should therefore be read as a self-consistent approximation of the SCIN aggregation rather than a literal replication of it.

The mapping from free-text model outputs to SCIN categories has asymmetric coverage: 100\% for GPT-4o-mini and Gemini 3 Flash Preview but only 89.4\% for Claude Sonnet 4.6, with any unmapped prediction treated as incorrect on every metric. We estimate that this depresses the absolute Q4 number for Claude Sonnet 4.6. We expect that the rank order across the three models is robust to this asymmetry, but the absolute gap between Claude Sonnet 4.6 and Gemini 3 Flash Preview is likely overstated as a result.

All four of our agreement metrics require exact post-mapping label equality with the dermatologist labels and do not reward clinical near-misses. The 370-category SCIN label space contains fine-grained splits such as the four separate categories for Tinea Corporis, Tinea Versicolor, Tinea Cruris, and Tinea Pedis, which differ only by body site. A model that correctly identifies a case as a superficial fungal infection but selects the wrong body-site variant therefore scores zero on Q2, Q3, and Q4, even though a clinician would consider the differential largely correct. All four of our numbers should therefore be read as a conservative floor on clinical usefulness.

Finally, the three models we evaluate all sit at the low to mid tier in terms of pricing. The top-tier flagship models from the same three providers, including GPT-5.4, Claude Opus 4.6, and Gemini 3.1 Pro, were not evaluated due to access and cost reasons.

\section{Discussion \& Conclusions}

The headline finding of our evaluation is that the best of the three tested models is essentially equal to an individual dermatologist on the strictest of our four agreement metrics, and does so at very low inference cost. On Q4, the exact top-1 consensus metric, the individual-dermatologist baseline is 28.6\% and Gemini 3 Flash Preview under the images-only condition A reaches 29.1\%. On Q3, the top-$k$ double-peer-support metric, both the baseline and the model reach 30.8\% on our evaluation subset. What makes this notable is the cost at which it is obtained. At the time of writing, a single call to Gemini 3 Flash Preview on a SCIN case costs approximately \$0.0045 on average, which is less than half a US cent per case. A commercially available generalist vision language model, with no task-specific training, therefore reaches the level of an individual dermatologist on the strictest of our agreement metrics at a per-case inference cost of well under one US cent. We note that this parity only holds on the strictest metrics: on the more relaxed Q1 and Q2 metrics the best model trails the dermatologist baseline by several percentage points, and the statement that vision language models are close to an individual dermatologist on this task is therefore sensitive to which metric is chosen.

A second finding, with direct clinical relevance, concerns the calibration of the models' self-reported confidence. Across all three models, self-reported confidence is miscalibrated relative to how often the model's prediction is listed by at least one of the dermatologists reviewing the case, and more consequentially, the top-1 confidence is nearly constant across cases and at most weakly correlated with case difficulty measured by mean dermatologist confidence. A triage workflow that uses model-reported confidence in a threshold-based gate for referral will therefore fail on precisely the cases where the dermatologists themselves were uncertain, since the model reports the same high confidence on those cases as on any other. One caveat should be noted on this finding. The notion of calibration used here is defined relative to how often the model's prediction matches a dermatologist's, and as the agreement metrics in Section~\ref{sec:primary} make clear, dermatologists themselves often disagree with one another on these cases. The result should therefore not be read as a claim that the models are overconfident in some absolute sense, but rather that their confidence does not track empirical dermatologist agreement and does not distinguish cases the panel found easy from those it found difficult. The broader conclusion that confidence gating should not be relied upon in a clinical setting is robust to this caveat: even if the dermatologist panel is treated as a noisy reference rather than a perfect one, a model whose confidence is largely independent of panel difficulty remains unsafe to use as a gating signal.

Two lesser observations complete the picture. First, within the mid-tier price range we tested, model cost is not predictive of performance on this task. Gemini 3 Flash Preview is both the cheapest and the strongest of the three, with a consistent Gemini 3 Flash $>$ Claude Sonnet 4.6 $>$ GPT-4o-mini rank order across all four metrics and both prompt conditions. This ordering is specific to the three models we tested and may not extend above the mid-tier band. Second, the effect of supplying additional patient metadata to the models is inconclusive across our small sample of three. Claude Sonnet 4.6 improves on all four metrics when the metadata is supplied in condition B, Gemini 3 Flash Preview regresses on all four, and GPT-4o-mini improves on three of the four, so no general statement about the value of patient context for this class of model can be drawn. What is notable is that one of the three models degrades on every metric when metadata is added, which runs counter to the intuitive expectation that additional information about the patient should help a model narrow its differential.

In summary, we compared three recently released vision language models against individual held-out dermatologists on a 78-case subset of the SCIN dataset, using a leave-one-out protocol in which the same indicator function and the same peer reference are applied to both the model and the dermatologist at each held-out position. Our main result is that the best of the three models is essentially equal to an individual dermatologist on the strictest of our four agreement metrics, at a per-case inference cost of less than half a US cent at the time of writing. The second main result is that none of the three models' self-reported confidences are safe to use as a triage signal, because their top-1 confidence is effectively flat across cases and at most weakly tracks case difficulty. In addition to the main findings, we also observed that model cost was not predictive of performance within the mid-tier range we tested, and the effect of supplying patient metadata was inconclusive across models but counterintuitive in direction for one of the three. The obvious extension to this study is to evaluate the same protocol on the top-tier flagship models, which we leave to future work.

%
%
\bibliographystyle{splncs04}
\bibliography{references}

\end{document}